\pdfoutput=1

\documentclass[11pt]{article}

\usepackage{acl}

\usepackage{times}
\usepackage{latexsym}

\usepackage[T1]{fontenc}

\usepackage[utf8]{inputenc}

\usepackage{microtype}

\usepackage{amsmath}
\usepackage{amsfonts}
\usepackage{subfigure}
\usepackage{graphicx}
\usepackage{booktabs}
\usepackage{url}
\usepackage{tabularx}
\usepackage{bigstrut}
\usepackage{multirow}
\usepackage{color}
\usepackage{xcolor}

\usepackage{verbatimbox}

\usepackage{enumitem}
\usepackage{xspace}

\newcommand{\MODEL}{LASER\xspace}
\newcommand{\smallsection}[1]{\noindent\textbf{#1.}}

\newcommand{\revise}[1]{{#1}}

\newcommand{\delete}[1]{\textcolor{gray}{}}

\newcommand\blfootnote[1]{%
  \begingroup
  \renewcommand\thefootnote{}\footnote{#1}%
  \addtocounter{footnote}{-1}%
  \endgroup
}

\title{Towards Few-shot Entity Recognition in Document Images: \\A Label-aware Sequence-to-Sequence Framework}

\author{Zilong Wang \\
  University of California, San Diego \\
  \texttt{zlwang@ucsd.edu} \\\And
  Jingbo Shang$^{*}$ \\
  University of California, San Diego \\
  \texttt{jshang@ucsd.edu} \\}

\begin{document}
\maketitle
\begin{abstract}
\blfootnote{$*$ Jingbo Shang is the corresponding author.}
Entity recognition is a fundamental task in understanding document images.
Traditional sequence labeling frameworks \revise{treat the entity types as class IDs and rely on extensive data and high-quality annotations to learn semantics which are typically expensive in practice.}
In this paper, we aim to build an entity recognition model requiring only a few shots of annotated document images.
To overcome the data limitation, we propose to leverage the label surface names to better inform the model of the target entity type semantics
\revise{and also embed the labels into the spatial embedding space to capture the spatial correspondence between regions and labels}. 
Specifically, we go beyond sequence labeling and develop a novel label-aware seq2seq framework, \MODEL.
\revise{The proposed model follows a new labeling scheme that} generates the label surface names word-by-word explicitly after generating the entities.
During training, \MODEL refines the label semantics by updating the label surface name representations and also strengthens 
the label-region correlation.
In this way, \MODEL recognizes the entities from document images through both semantic and layout correspondence.
Extensive experiments on two benchmark datasets demonstrate the superiority of \MODEL under the few-shot setting.

\end{abstract}

\section{Introduction}

Entity recognition lies in the foundation of document image understandings, which aims at extracting
word spans that perform certain roles from the document images, such as \textit{header}, \textit{question}.
Distinct from the text-only named entity recognition task, the document images, such as forms, tables, receipts, and multi-columns, provide a perfect scenario to apply multi-modal techniques into practice where the rich layout formats in such document images serve as the new, complementary signals for entity recognition performance in addition to the existing textual data.

Recent methods~\citep{xu2020layoutlm,hong2020bros,garncarek2021lambert} follow the traditional sequence labeling framework to extract the word spans using the standard \texttt{IOBES} tagging schemes~\citep{marquez2005semantic,ratinov2009design} in named entity recognition tasks. 
\revise{Entity types are treated as class IDs and the semantics of the label surface names are ignored.} These methods also largely extend the label space by including combinations of the boundary identifiers (\texttt{B}, \texttt{I}, \texttt{E}, \texttt{S}) and entity types.
For instance, when there are 3 target entity types, the extended label space would have $13$ (i.e., $4 \times 3 + 1$) dimensions.
As a result, they \revise{fail to learn from the data efficiently and} require extensive datasets and high-quality annotations to create the connection between entities and their entity types.
\revise{Meanwhile,} document images typically include various formats and have a high \revise{diversity} of entities within each page. \revise{It is expensive or almost impossible to enumerate all required entity types and obtain enough annotated data for them.}
Moreover, ethical concerns would arise when it comes to the receipts or consent forms, which makes it even harder to collect enough data.

\revise{Due to the inefficiency of traditional methods and the data limitation in real application scenarios, it is necessary to resort to few-shot learning for entity recognition in document images. We aim at exploiting the potential of a limited number of training pages and try to generalize our model on the much larger number of new pages for testing.}
In our method, we go beyond the sequence labeling framework and reformulate the entity recognition as a sequence-to-sequence task. 
Specifically, we propose a new generative labeling scheme for entity recognition --- the label surface name is explicitly generated right after each entity as a part of the target sequence.
In this way, different entity types are no longer independent dimensions in the label space \revise{and models can leverage the semantic connect between the entities and entity types.}

To this end, we propose a \underline{l}abel-\underline{a}ware \underline{s}equence-to-sequence framework for \underline{e}ntity \underline{r}ecognition, \MODEL. \revise{Our implementation is based on pre-trained language model LayoutReader~\footnote{Licensed under the MIT License}~\citep{wang2021layoutreader}, which is a layout-aware pre-trained sequence-to-sequence model.}

\begin{figure*}[t]
    \centering
    \includegraphics[width=\linewidth]{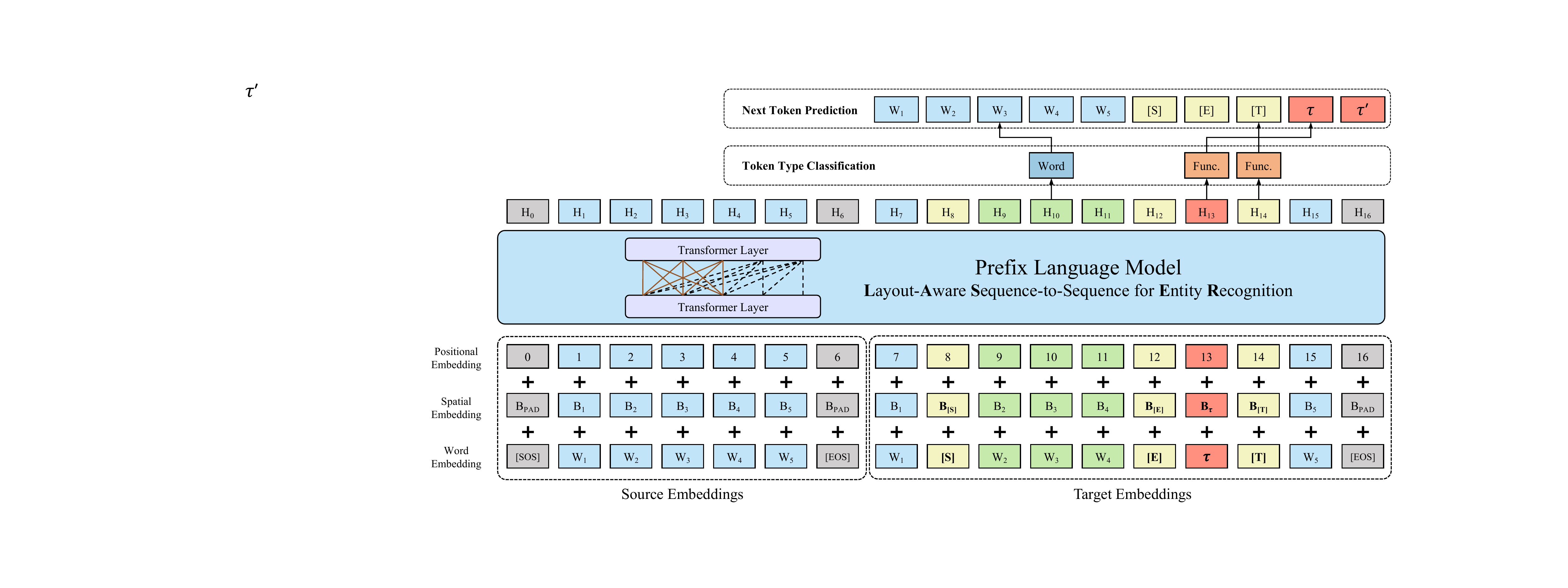}
    \caption{The Framework of \MODEL: \texttt{[B]}, \texttt{[E]}, \texttt{[T]} denote the boundaries; $\tau$, $\tau^\prime$, $\tau^{\prime\prime}$ are the label surface names; (a) is the process of generative labeling scheme; (b) shows the alignment of the spatial identifiers and embeddings.
    }
    \label{fig:model}
\end{figure*}

As shown in Figure~\ref{fig:model}, \MODEL extends the architecture of LayoutReader for our proposed generative labeling scheme to better solve the few-shot entity recognition task for document images.
Specifically, after generating certain word spans, the model can choose to generate either the following words in the source sequence or label surface names. \revise{The entity labels are explicitly inserted in the generated sequence so that the probability of the entity types conditioned on the entity, ${P}(\mbox{type}|\mbox{entity})$, can be maximized not only by the signals from the training data but also by the knowledge from the pre-training of the language models.} \revise{We also embed the label surface names into the spatial embedding space,} so the \revise{generation of labels} is also aware of the correlation between labels and the regions in the page.


\revise{Benefit from the novel generative labeling scheme and the semantics of labels}, \MODEL is able to effectively recognize entities in document images with only a limited number of training samples. 
\revise{In contrast, the sequence labeling models use less efficient tagging scheme, thus requiring more data and failing in the few-shot settings.}

We validate \MODEL using two benchmarks, FUNSD~\citep{jaume2019} and CORD-Lv1~\citep{park2019cord}. 
Both datasets are from real scenarios and fully-annotated with textual contents and bounding boxes.
We compare our model with strong baselines and study the label-entity semantic and spatial correlations. 
We summarize our contribution as follows.
\begin{itemize}[leftmargin=*,nosep]
    \item We reformulate the entity recognition task and propose a new generative labeling scheme that embeds the label surface names into the target sequence to explicitly inform the model of the label semantics.
    \item We propose a novel label-aware sequence-to-sequence framework \MODEL to better handle few-shot entity recognition tasks for document images than the traditional sequence labeling framework using both label semantics and layout format learning.
    \item Extensive experiments on two benchmark datasets demonstrate the effectiveness of \MODEL under few-shot settings.
\end{itemize}
\noindent\textbf{Reproducibility.} We will release the code and datasets on Github\footnote{\url{github.com/zlwang-cs/LASER-release}}.

\section{Problem Formulation}
The few-shot entity recognition in the document images is to take the text and layout inputs from a limited number of training samples to predict the boundary of each entity and classify the entity into categories. 
Given a document image page $\mathcal{P}$, the words within the page are annotated with their textual contents $w$ and the bounding boxes $B=(x_0, y_0, x_1, y_1)$ (top-left and bottom-right corners) \revise{by human annotators or the OCR engines, and all the words and bounding boxes are listed in a sequence
serving as the inputs from textual and layout modalities.} In this way, the entities are spans of these words referring to precise concepts, which makes it possible to conduct entity recognition using sequence labeling or generative labeling scheme.
We randomly select a small subset of training samples and evaluate the performance under the $k$-shot training, where $k$ denotes the number of the training samples.

\section{Our Generative Labeling Scheme}

We propose our labeling scheme of entity recognition in the generative manner which generates the entity boundaries and the label surface names explicitly.
Specifically, given an entity $e=[w_i,w_{i+1}...,w_{j}]$, we use the \texttt{[B]} and \texttt{[E]} to denote the boundary of the entity and append the label surface name afterwards. Overall, the generative formulation is to generate:
\begin{align*}
    w_{i-1},\textbf{\texttt{[B]}},w_i,...,w_{j},\textbf{\texttt{[E]}},\mathbf{\tau_1},...,\mathbf{\tau_k},\textbf{\texttt{[T]}},w_{j+1}
\end{align*}
where \texttt{[B]} and \texttt{[E]} denote the start and end of the entity; $\tau_1...\tau_k$ are the words in the label surface name; \texttt{[T]} denotes the end of the label surface name.
For example, ``\textit{Sender}'' and ``\textit{Charles Duggan}'' are a pair of \textit{question} and \textit{answer} from a document image. 
According to the generative labeling scheme, the corresponding generated sequence is that: \texttt{[B]} \textit{Sender} \texttt{[E]} \textit{question} \texttt{[T]} \texttt{[B]} \textit{Charles Duggan} \texttt{[E]} \textit{answer} \texttt{[T]}.

\section{Our \MODEL Framework}

In this section, we introduce our label-aware sequence-to-sequence framework for entity recognition in document images. 
First, we 
introduce our method in 
a bird's eye view.
Then we dive into the details of each part including the multi-modal prefix language model, the label-aware generation.

\subsection{Overview}

Our proposed \MODEL is a label-aware sequence-to-sequence model for entity recognition in document images. The framework is shown in Figure \ref{fig:model}.
The model follows the prefix language model paradigm~\citep{raffel2019exploring,unilm,unilmv2} and is built upon the pre-trained language model, LayoutReader~\citep{wang2021layoutreader}.
With extensive knowledge learned in pre-training stage, the model leverages the semantic meaning of label surface names during generation. 

Since the functional tokens (e.g. \texttt{[B]}, \texttt{[E]}) and the label surface names are foreign words in the given page, their layout features are nonexistent. We use trainable vectors as special layout identifiers for these extra tokens and these vectors are well aligned into the spatial embedding space. In this way, the spatial correspondence between layout formats and labels can be learned.

To reinforce the model to distinguish the functional tokens (e.g. \texttt{[B]}, \texttt{[E]}) and ordinary words, an extra binary classification module is added, and the probability is used in the next token prediction. 

Equipped with all the components, our proposed model is able to conduct entity recognition efficiently and effectively under the few-shot setting.

\subsection{Multi-modal Prefix LM}\label{sec:mmplm}

\MODEL is built on the layout-aware prefix language model, LayoutReader~\citep{wang2021layoutreader}. 
Prefix language model refers to a multi-layered Transformer where the source sequence and target sequence are packed together and a ``partially-triangle'' mask is used to control the attention between tokens in the two sequences. 
In \MODEL, the source sequence has full self-attention and the target sequence only attends to the previous tokens so the conditional generative probability is learned.

\paragraph{Input Embedding} The input embedding layer of \MODEL includes the word embedding, spatial embedding, and positional embedding. 
We normalize and round the bounding box coordinates to integers ranging from 0 to 1000, and embed them as trainable vectors as spatial embeddings~\citep{xu2020layoutlm,Xu2020LayoutLMv2MP,Xu2020LayoutXLMMP,wang2021layoutreader}.
So the input embeddings of the ordinary words are as follows:
\begin{align*}
\small
    e_{w_i} = \mbox{WordEmb}(w_i) + \mbox{SpatialEmb}(B_i) + \mbox{PosEmb}(i)
\end{align*}
where WordEmb, SpatialEmb, PosEmb are the word embedding, the spatial embedding, and the positional embedding lookup tables, respectively; $i$ is the index of the word in the packed sequence.

The functional tokens and label surface names are new tokens in the given page. We cannot extract the layout features from the bounding boxes of them because their bounding boxes are nonexistent. Instead of the actual bounding boxes, we design unique embedding vectors for each new tokens as their layout identifiers. 
These identifiers can perform in the same way as real bounding boxes during training \revise{to embed the functional tokens and label surface names into the spatial embedding space.}
The input embedding replaces the spatial embedding with the spatial identifiers:
\begin{align*}
\small
    e_{\lambda} = \mbox{WordEmb}(\lambda) + \mbox{SpatialID}(\lambda) + \mbox{PosEmb}(i)
\end{align*}
where SpatialID is the spatial identifier lookup table; $i$ is the index of the word in the packed sequence; $\lambda \in \{\texttt{[B]},\texttt{[E]},\texttt{[T]},\tau_1,...,\tau_t\}$.

Within the input embedding layer, the pre-trained model learns the semantic and layout formats from word embeddings or spatial features. The spatial embeddings are already pre-trained and further fine-tuned in the downstream tasks, and the spatial identifiers are new to the model and completely trained in the downstream tasks.


\paragraph{Attention Mask} As mentioned, \MODEL depends on a ``partially-triangle'' mask to realize sequence-to-sequence training within one encoder. To be more specific, the ``partially-triangle'' attention mask has two parts, the source part and the target part.
In the source part, the tokens can attend to each other, which enables the model to be aware of the entire sequence. In the target part, to predict the next token in a sequence-to-sequence way, we design the ``triangle'' mask which prevents the tokens from attending to the tokens after them. Therefore, the generative probability conditioned on the previous tokens can be computed.

\paragraph{Output Hidden States}
To learn the conditional generative probability of the next token, we take the output hidden states corresponding to the target sequence which is denoted as $\mathbf{h}_{n+1},\mathbf{h}_{n+2},...,\mathbf{h}_{n+m}$, where $n+1$ is the beginning of the target sequence in the packed sequence. According to the ``partially-triangle'' attention mask, $\mathbf{h}_{n+k}$ is produced with the attention to the source tokens and the previous target tokens, i.e., the input embeddings whose index ranges from $1$ to $n+k$. Therefore, $\mathbf{h}_{n+k}$ is used to predict the $(k+1)$-th token in the target sequence.

\subsection{Label-aware Generation}
In the sequence-to-sequence setting, \MODEL estimates the probability of next token conditioned on the previous context, i.e. $P(x_{k}|x_{<k})$ and $x_k \in \mathcal{C}$, where $\mathcal{C}=\{w_1...w_n\}\cup \{\tau_1...\tau_t\}\cup \{\texttt{[B]},\texttt{[E]},\texttt{[T]}\}$ is the set of all candidate words.
Following LayoutReader, we restrain the candidates within the source words instead of the whole dictionary, and we go beyond it and extend the candidate set to include the functional tokens and label surface names. Moreover, to distinguish whether the next word belongs to the source or not, we design an extra binary classification module.

Specifically, we take the hidden states $h_{k}$ to predict whether the next token is from the source or not. We denote the probability $P(x_{k+1}\in \mbox{src}) = p_{k+1}$. Then we use $p_{k+1}$ to weight the next token prediction. The probability that the next token is the $i$-th word in the source is computed as follows:
\begin{align*}
    P(x_{k+1}=w_i|x_{\le k}) = \frac{p_{k+1} \exp{\left(\mathbf{e}_{w_i}^T\mathbf{h}_{k}+{b}_{k}\right)}}{\sum_j\exp{\left(\mathbf{e}_{w_j}^T\mathbf{h}_{k}+{b}_{k}\right)}}
\end{align*}

where $w_i$ is the $i$-th word in the source; $\mathbf{e}_{w_i}$ is the input embedding of $w_i$; $\mathbf{b}_k$ is the bias.

Similarly, the probability that the next token is one of the functional tokens or label surface names is computed as follows:
\begin{align*}
    P(x_{k+1}=\lambda|x_{\le k}) = \frac{(1-p_{k+1})\exp{\left(\mathbf{e}_{\lambda}^T\mathbf{h}_{k}+{b}^\prime_{k}\right)}}{\sum_{\lambda^\prime} \exp{\left(\mathbf{e}_{\lambda^\prime}^T\mathbf{h}_{k}+{b}^\prime_{k}\right)}}
\end{align*}
where $\lambda$ is a functional token or label surface name, i.e. $\lambda \in \{\texttt{[B]},\texttt{[E]},\texttt{[T]},\tau_1,...,\tau_t\}$; $1-p_{k+1}$ is the probability that $(k+1)$-th token is a functional token or label surface name; $\mathbf{b}^\prime_k$ is the bias.

\paragraph{Label Semantics Learning}

With the log likelihood loss of generative language modeling, the model maximize the dot production between the hidden states $\mathbf{h}$ and the input embeddings $e$. The semantic correlation is learned considering that the input embeddings of the labels surface names are encoded in the word embeddings.

\paragraph{Spatial Identifier Learning}

From the layout format perspective, the input embedding of the label surface names also includes the spatial identifiers. 
When predicting the next token, the log likelihood also strengthens the relation between the spatial identifiers and the layout context. 
In this way, \MODEL inserts the spatial identifiers into the hyperspace of the spatial embeddings. In other words, \MODEL predicts where a certain label is more likely to be.
Similar to the joint probability of language modeling, \MODEL maximizes the joint probability of a mixture of spatial identifiers and spatial embeddings: $P(...,B_{k-1},B_{k},\tau,B_{k+1},...)$
where $B_k$ is the bounding boxes of the words in the page and the $\tau$ is the label to predict. Further visualization is conducted in Section \ref{sec:vis}.

\subsection{Sequential Decoding}
After training, \MODEL follows the prefix language modeling paradigm and generates the target sequence sequentially. We input the source sequence into the model and take the last hidden states to predict the first token in the target. Then we append the result to the end of input and repeatedly run the generation. We cache the states of the model and achieve generation in linear time.

\section{Experiments}


In this section, we conduct experiments and ablation study on FUNSD~\citep{jaume2019} and CORD-Lv1~\cite{park2019cord} under few-shot settings. We replace the original label surface names with other tokens to study the importance of semantic meaning. We also plot the heatmaps of the similarity between the spatial identifiers and the spatial embeddings to interpret the spatial correspondence. Case studies are also conducted.

\subsection{Experimental Setups}
\revise{All the experiments are under few-shot settings using 1, 2, 3, 4, 5, 6, 7 shots. We use 6 different random seeds to select the few-shot training samples and the data augmentation is conducted to solve the data sparsity. We train all the models using the same data and compute the average performance and the standard deviation. We only report the result of 1, 3, 5, 7 shots for space limitation.} To evaluate our model, we first convert our results into \texttt{IOBES} tagging style and compute the word-level precision, recall, and F-1 score using the APIs from \citet{seqeval} so that all comparisons with sequence labeling methods are under the same metrics. We believe such experiment settings guarantee the results are representative.

\subsection{Datasets}
Our experiments are conducted on two real-world data collections: FUNSD and CORD-Lv1. Both datasets provide rich annotations for the document image understandings includes the words and the word-level bounding boxes. The details and statistics of these two datasets are as follows.
\begin{itemize}[nosep, leftmargin=*]
    \item \textbf{FUNSD:} FUNSD consists of 199 fully-annotated, noisy-scanned forms with various appearance and format which makes the form understanding task more challenging. The word spans in this datasets are annotated with three different labels: \texttt{header}, \texttt{question} and \texttt{answer}, and the rest words are annotated as \texttt{other}. We use the original label names.
    \item \textbf{CORD-Lv1:} CORD consists of about 1000 receipts with annotations of bounding boxes and textual contents. The entities have multi-level labels. We select the first level and denote the dataset as CORD-Lv1. The first level includes \texttt{menu}, \texttt{void-menu}, \texttt{subtotal} and \texttt{total}. We simplify \texttt{subtotal} as \texttt{sub} and \texttt{void-menu} as \texttt{void}.
\end{itemize}

\begin{table}[t]
    \centering
    \caption{Dataset Statistics.}
    \label{tab:addlabel}%
    \small
  \resizebox{\linewidth}{!}{
    \setlength{\tabcolsep}{0.8mm}{
\begin{tabular}{lccc}
\toprule
\textbf{Dataset} & \textbf{\# Train Pages} & \textbf{\# Test Pages} & \textbf{\# Entities / Page} \\
\midrule
FUNSD & 149   & 50    & 42.86 \\
CORD-Lv1 & 800   & 100   & 13.82 \\
\bottomrule
\end{tabular}%
}
}
\end{table}%

\begin{table*}[t]
    \centering
    \caption{Evaluation Results with Different Sizes of Few-shot Training Samples: \textbf{Bold} denotes the best model; \underline{Underline} denotes the second-best model. 
    }
    \label{tab:main}%
\small
\resizebox{\linewidth}{!}{
    \setlength{\tabcolsep}{3mm}{
\begin{tabular}{clcccccc}
\toprule
\multirow{2}[4]{*}{$|\mathcal{P}|$} & \multicolumn{1}{l}{\multirow{2}[4]{*}{\textbf{Model}}} & \multicolumn{3}{c}{\textbf{FUNSD}} & \multicolumn{3}{c}{\textbf{CORD-Lv1}} \\
\cmidrule(lr){3-5} \cmidrule(lr){6-8}     &       & \textbf{Precision} & \textbf{Recall} & \textbf{F-1} & \textbf{Precision} & \textbf{Recall} & \textbf{F-1} \\
\midrule
\multirow{5}[2]{*}{1} & BERT  & 9.62$\pm$2.24 & 24.14$\pm$3.46 & 13.55$\pm$2.09 & 30.64$\pm$2.80 & 45.60$\pm$3.45 & 36.64$\pm$3.10 \\
      & RoBERTa & 9.29$\pm$1.57 & 22.06$\pm$5.64 & 12.76$\pm$1.91 & 30.66$\pm$4.25 & 44.39$\pm$6.72 & 36.25$\pm$5.18 \\
      & LayoutLM & \underline{11.39$\pm$1.12} & \underline{24.73$\pm$7.38} & \underline{15.18$\pm$2.17} & \underline{33.27$\pm$7.32} & \textbf{49.49$\pm$10.26} & \underline{39.77$\pm$8.47} \\
      & LayoutReader & 11.32$\pm$0.62 & 22.53$\pm$4.80 & 14.84$\pm$1.25 & 32.17$\pm$4.64 & \underline{45.61$\pm$6.54} & 37.70$\pm$5.31 \\
      & LASER & \textbf{30.40$\pm$4.89} & \textbf{35.20$\pm$7.20} & \textbf{32.36$\pm$5.14} & \textbf{47.63$\pm$3.90} & 45.52$\pm$5.84 & \textbf{46.24$\pm$3.01} \\
\midrule
\multirow{5}[2]{*}{3} & BERT  & 16.42$\pm$4.30 & 34.74$\pm$5.36 & 22.19$\pm$5.05 & 39.62$\pm$3.99 & 56.65$\pm$4.03 & 46.58$\pm$3.94 \\
      & RoBERTa & 16.71$\pm$3.63 & 31.28$\pm$3.55 & 21.66$\pm$3.84 & 44.51$\pm$4.69 & 60.18$\pm$4.69 & 51.15$\pm$4.70 \\
      & LayoutLM & \underline{28.67$\pm$6.56} & \textbf{47.22$\pm$8.31} & \underline{35.42$\pm$7.00} & \underline{47.68$\pm$7.49} & \textbf{63.93$\pm$7.04} & \underline{54.57$\pm$7.46} \\
      & LayoutReader & 22.37$\pm$2.03 & 35.19$\pm$4.97 & 27.19$\pm$2.56 & 43.85$\pm$4.72 & 56.90$\pm$2.47 & 49.47$\pm$3.95 \\
      & LASER & \textbf{43.66$\pm$1.97} & \underline{47.08$\pm$5.72} & \textbf{45.21$\pm$3.74} & \textbf{61.16$\pm$3.11} & \underline{60.33$\pm$5.65} & \textbf{60.63$\pm$4.00} \\
\midrule
\multirow{5}[2]{*}{5} & BERT  & 20.57$\pm$2.59 & 39.25$\pm$1.10 & 26.93$\pm$2.46 & 45.73$\pm$4.31 & 63.29$\pm$3.68 & 53.06$\pm$4.14 \\
      & RoBERTa & 19.47$\pm$2.32 & 35.04$\pm$1.89 & 24.94$\pm$1.93 & 52.21$\pm$4.55 & \underline{66.63$\pm$5.52} & 58.54$\pm$4.92 \\
      & LayoutLM & \underline{39.24$\pm$4.33} & \textbf{58.20$\pm$2.45} & \underline{46.72$\pm$3.12} & \underline{56.13$\pm$7.39} & \textbf{71.66$\pm$6.13} & \underline{62.91$\pm$7.04} \\
      & LayoutReader & 27.52$\pm$3.44 & 41.17$\pm$4.01 & 32.89$\pm$3.28 & 51.97$\pm$8.42 & 63.82$\pm$7.87 & 57.24$\pm$8.32 \\
      & LASER & \textbf{47.25$\pm$1.93} & \underline{52.85$\pm$1.22} & \textbf{49.87$\pm$1.29} & \textbf{65.62$\pm$3.79} & 64.90$\pm$5.78 & \textbf{65.23$\pm$4.70} \\
\midrule
\multirow{5}[2]{*}{7} & BERT  & 21.44$\pm$2.07 & 40.87$\pm$3.79 & 28.09$\pm$2.48 & 50.13$\pm$4.35 & 66.67$\pm$3.67 & 57.20$\pm$4.07 \\
      & RoBERTa & 23.68$\pm$3.06 & 38.74$\pm$3.54 & 29.32$\pm$3.08 & 55.14$\pm$4.49 & \underline{69.35$\pm$4.16} & 61.43$\pm$4.42 \\
      & LayoutLM & \underline{43.23$\pm$5.27} & \textbf{61.73$\pm$5.97} & \underline{50.76$\pm$5.30} & \underline{62.87$\pm$3.98} & \textbf{76.38$\pm$2.72} & \textbf{68.96$\pm$3.49} \\
      & LayoutReader & 31.22$\pm$3.14 & 45.08$\pm$3.83 & 36.85$\pm$3.26 & 54.43$\pm$5.89 & 65.48$\pm$5.34 & 59.42$\pm$5.68 \\
      & LASER & \textbf{50.62$\pm$3.26} & \underline{53.63$\pm$2.89} & \textbf{51.98$\pm$2.00} & \textbf{68.02$\pm$3.16} & 66.87$\pm$4.82 & \underline{67.40$\pm$3.76} \\
\bottomrule
\end{tabular}%
}
}
\end{table*}%

\subsection{Compared Methods}

We evaluate \MODEL against several strong sequence labeling methods as follows. 
\begin{itemize}[nosep, leftmargin=*]
    \item \textbf{BERT}~\citep{devlin2018bert} is a text-only auto-encoding pre-trained language model using the large-scale mask language modeling. We fine-tune the pre-trained BERT-base model with the few-shot training samples on each datasets.
    \item \textbf{RoBERTa}~\citep{liu2019roberta} extends the capacity of BERT and achieves better performance in multiple natural language understanding tasks. We also conduct the fine-tuning with few-shot training samples.
    \item \textbf{LayoutLM}~\citep{xu2020layoutlm} is a multi-modal language model which includes the layout and text information. It is built upon BERT and adds the extra spatial embeddings into the BERT embedding layer. Following LayoutLM, LayoutLMv2~\citep{Xu2020LayoutLMv2MP} leverages extra computer vision features and improves the performance, which are strong signals but absent in our settings. For a fair comparison, we do not include LayoutLMv2 in our comparative experiments.
    \item \revise{\textbf{LayoutReader}~\citep{wang2021layoutreader} is a layout-aware sequence-to-sequence model for reading order detection. We append a linear layer upon the hidden states to conduct sequence labeling.}
\end{itemize}
These compared methods are in their base version and follow the \texttt{IOBES} tagging scheme.
\delete{We denote our model as {\MODEL}.
In \MODEL, we argue that the semantic meaning of label surface names helps the model learn the semantic correspondence between labels and entities when there are not enough training samples under few-shot learning settings. 
In order to understand how much the label surface names can tell, we introduce an ablation version {\MODEL (IRLVT)} by replacing the label surface names with irrelevant tokens, which by default are [\textit{w}, \textit{x}, \textit{y}, \textit{z}]. }


\begin{figure}[t] 
\centering
  \subfigure[FUNSD]{%
    \includegraphics[width=.48\linewidth]{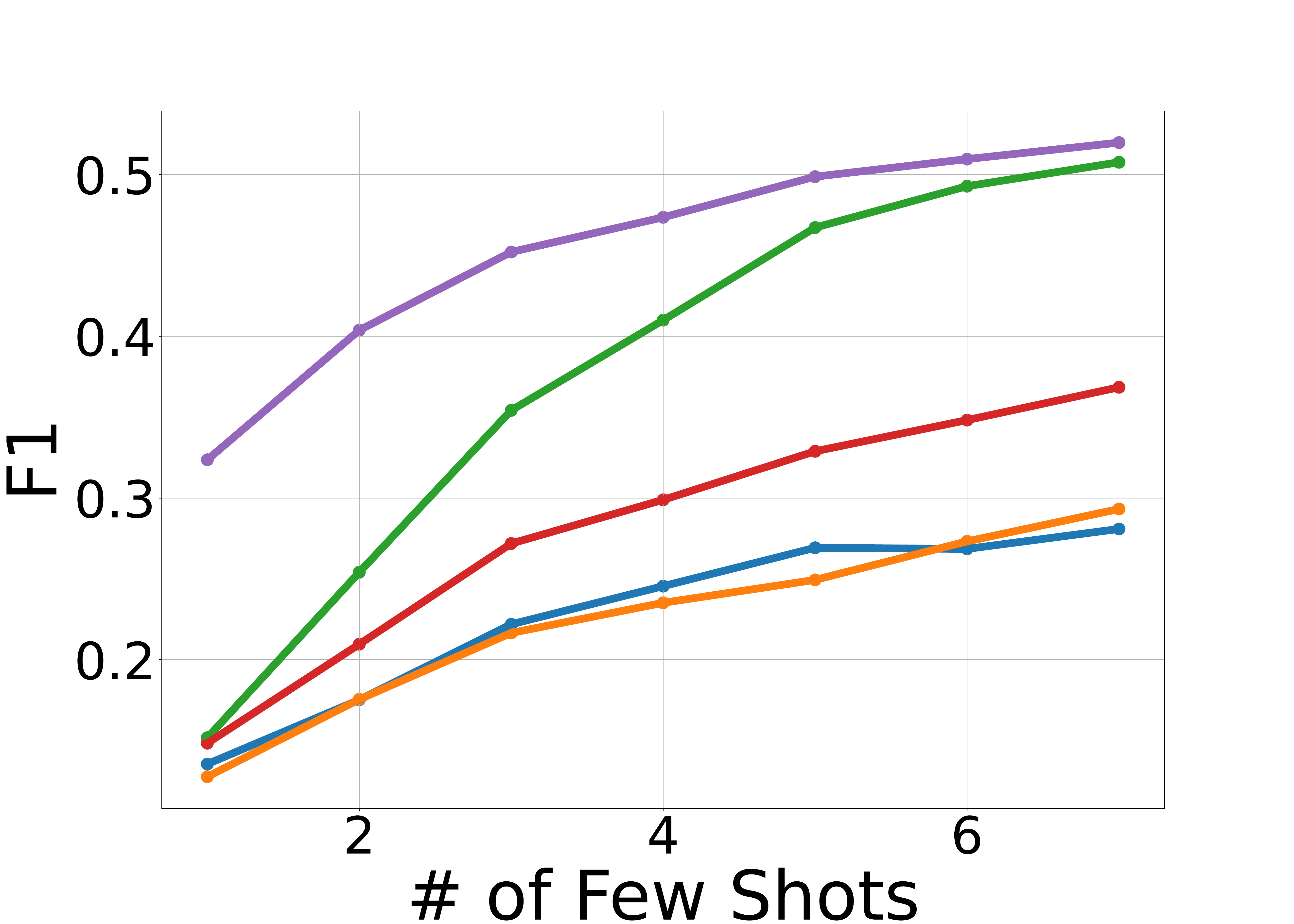}
  } 
  \hfill 
  \subfigure[CORD-Lv1]{%
    \includegraphics[width=.48\linewidth]{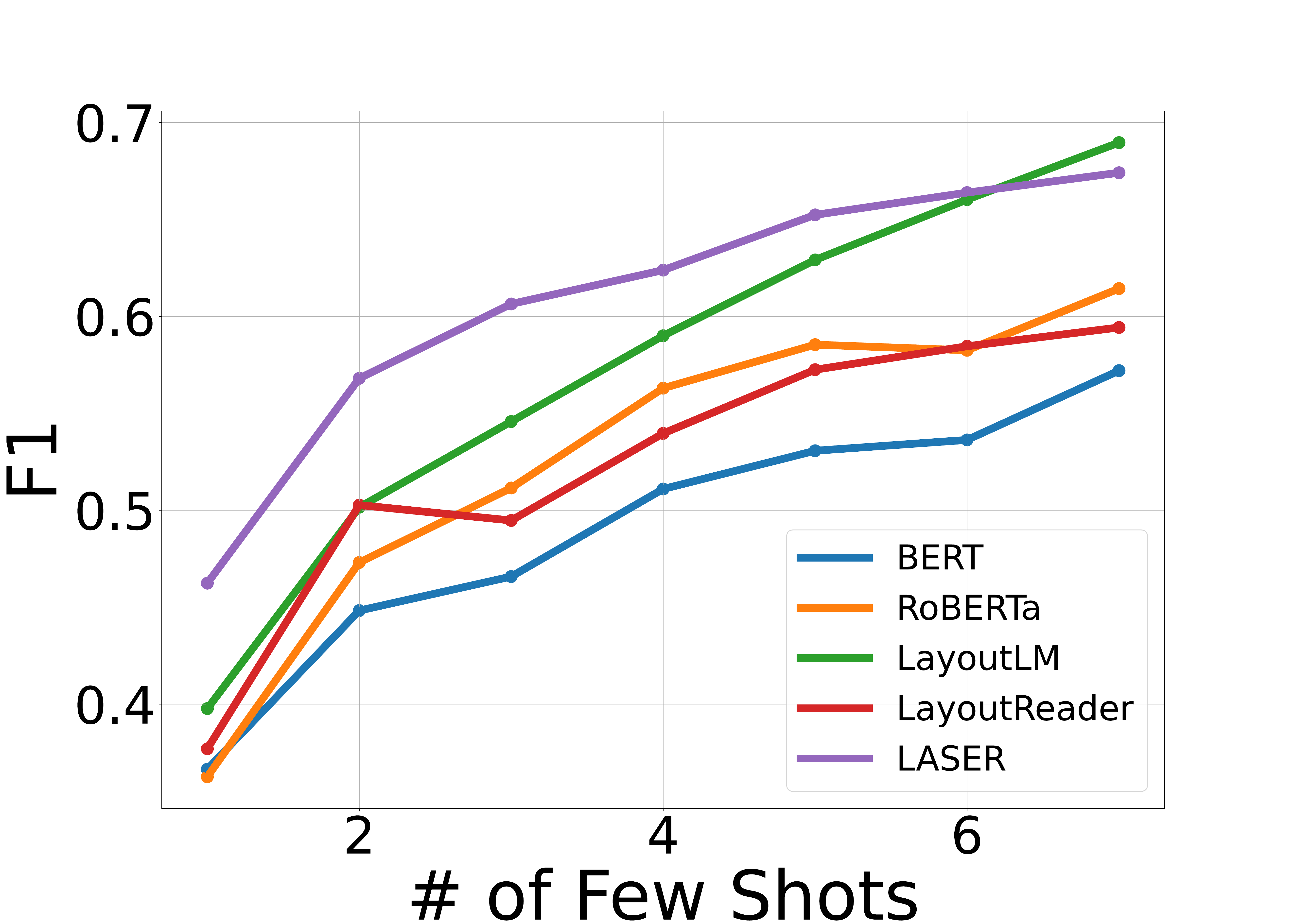}
  } 
  \caption{F-1 Curves with Different Sizes of Few-shot Training Samples.} 
  \label{fig:curve}
\end{figure}

\subsection{Implementation Details}

We build \MODEL on the base of LayoutReader. We use the Transformers~\citep{wolf2019huggingface} and the s2s-ft toolkits from the repository of \citet{unilm}. We use one NVIDIA A6000 to finetune with batch size of 8. We optimize the model with AdamW optimizer and the learning rate is $5\times10^{-5}$. 

\subsection{Experimental Results}


From Table~\ref{tab:main} and Figure~\ref{fig:curve}, the results show that, under few-shot settings, our proposed model, \MODEL, achieves the SOTA overall performance compared with sequence labeling models. \revise{We conclude that the gain of performance comes mostly from the generative labeling scheme since \MODEL largely outperforms LayoutReader although both of them share the same backbone.}

Specifically, compared with the second-best baseline, \MODEL improves the F-1 scores by 8.59\% on FUNSD and by 3.32\% on CORD-Lv1 on average across the different shots and \MODEL (IRLVT) also surpasses the baselines under most settings.

Moreover, the improvement on precision is remarkable. \MODEL improves the precision by 12.35\% on FUNSD and by 10.62\% on CORD-Lv1 on average across the different shots. Especially, under 1-shot setting, it surpasses the best sequence labeling model on FUNSD by 19.01\% on precision, 10.47\% on recall and 17.18\% on F-1 score. 

We can also observe a drop in the improvement with the increasing number of training samples. 
We conclude that, with enough training samples, the sequence labeling learns the meaning of each label and the semantics of each label surface names no longer provides extra useful information. 

\revise{Based on these comparison, we safely come to the conclusion that our proposed generative labeling scheme is superior to the traditional sequence labeling scheme in few shot settings.}

\begin{table*}[t]
  \centering
  \caption{Ablation Study of Different Label Surface Names in \MODEL. {IRLVT} uses the irrelevant tokens as labels; {ORIG} uses the original label surface names; {Sub1} and {Sub2} use some reasonable alternative label surface names. as substitutes. 
    \textbf{Bold} denotes the best model; \underline{Underline} denotes the second-best model.}
  \label{tab:unk}%
  \small
  \resizebox{\linewidth}{!}{
    \setlength{\tabcolsep}{1.2mm}{
    \begin{tabular}{c lccc lccc}
        \toprule
        \multirow{2}[4]{*}{$|\mathcal{P}|$} & \multicolumn{4}{c}{\textbf{FUNSD}} & \multicolumn{4}{c}{\textbf{CORD-Lv1}} \\
        \cmidrule(lr){2-5} \cmidrule(lr){6-9}
        &   \textbf{Label Surface Names}    & \textbf{Precision} & \textbf{Recall} & \textbf{F-1} & \textbf{Label Surface Names} & \textbf{Precision} & \textbf{Recall} & \textbf{F-1} \\
        \midrule
        \multirow{4}[4]{*}{1} & IRLVT [\textit{x}, \textit{y}, \textit{z}] & 30.64$\pm$5.89 & 33.45$\pm$9.14 & 31.62$\pm$6.61 & IRLVT [\textit{w}, \textit{x}, \textit{y}, \textit{z}] & \textbf{48.57$\pm$4.93} & 44.12$\pm$6.36 & 45.84$\pm$3.57 \\
              & ORIG [\textit{header}, \textit{question}, \textit{answer}] & 30.40$\pm$4.89 & \underline{35.20$\pm$7.20} & 32.36$\pm$5.14 & ORIG [\textit{menu}, \textit{void}, \textit{sub}, \textit{total}] & 47.63$\pm$3.90 & \underline{45.52$\pm$5.84} & \underline{46.24$\pm$3.01} \\
        \cmidrule{2-9}
              & Sub1 [\textit{title}, \textit{key}, \textit{value}] & \textbf{31.78$\pm$4.75} & 34.21$\pm$7.44 & \underline{32.66$\pm$5.10} & Sub1 [\textit{info}, \textit{etc}, \textit{small}, \textit{number}] & \underline{48.12$\pm$4.15} & \textbf{48.47$\pm$6.60} & \textbf{48.04$\pm$4.06} \\
              & Sub2 [\textit{page}, \textit{topic}, \textit{value}]  & \underline{30.90$\pm$5.20} & \textbf{35.97$\pm$8.57} & \textbf{33.03$\pm$6.31} & Sub2 [\textit{page}, \textit{non}, \textit{part}, \textit{price}] & 45.59$\pm$5.68 & 44.09$\pm$7.87 & 44.38$\pm$5.39 \\
        \midrule
        \multirow{4}[4]{*}{3} & IRLVT [\textit{x}, \textit{y}, \textit{z}] & 43.51$\pm$1.46 & \underline{47.92$\pm$5.93} & \underline{45.44$\pm$3.36} & IRLVT [\textit{w}, \textit{x}, \textit{y}, \textit{z}] & 61.50$\pm$2.52 & 59.17$\pm$4.11 & 60.27$\pm$2.99\\
              & ORIG [\textit{header}, \textit{question}, \textit{answer}] & 43.66$\pm$1.97 & 47.08$\pm$5.72 & 45.21$\pm$3.74 & ORIG [\textit{menu}, \textit{void}, \textit{sub}, \textit{total}] & 61.16$\pm$3.11 & \textbf{60.33$\pm$5.65} & \underline{60.63$\pm$4.00}\\
        \cmidrule{2-9}
              & Sub1 [\textit{title}, \textit{key}, \textit{value}]  & \underline{43.87$\pm$1.33} & 47.11$\pm$6.07 & 45.26$\pm$3.44 & Sub1 [\textit{info}, \textit{etc}, \textit{small}, \textit{number}] & \underline{61.54$\pm$2.76} & 58.79$\pm$6.76 & 60.00$\pm$4.57\\
              & Sub2 [\textit{page}, \textit{topic}, \textit{value}]  & \textbf{43.88$\pm$1.34} & \textbf{48.01$\pm$6.86} & \textbf{45.65$\pm$3.93} & Sub2 [\textit{page}, \textit{non}, \textit{part}, \textit{price}] & \textbf{61.85$\pm$2.16} & \underline{60.29$\pm$2.85} & \textbf{61.03$\pm$2.10}\\
        \midrule
        \multirow{4}[4]{*}{5} & IRLVT [\textit{x}, \textit{y}, \textit{z}] & 46.94$\pm$1.87 & \underline{52.96$\pm$2.03} & 49.74$\pm$1.63 & IRLVT [\textit{w}, \textit{x}, \textit{y}, \textit{z}] & 63.67$\pm$3.82 & 61.10$\pm$5.21 & 62.33$\pm$4.48\\
              & ORIG [\textit{header}, \textit{question}, \textit{answer}] & 47.25$\pm$1.93 & 52.85$\pm$1.22 & \underline{49.87$\pm$1.29} & ORIG [\textit{menu}, \textit{void}, \textit{sub}, \textit{total}] & \textbf{65.62$\pm$3.79} & \textbf{64.90$\pm$5.78} & \textbf{65.23$\pm$4.70}\\
        \cmidrule{2-9}
              & Sub1 [\textit{title}, \textit{key}, \textit{value}]  & \underline{47.43$\pm$2.29} & 52.19$\pm$2.09 & 49.68$\pm$1.98 & Sub1 [\textit{info}, \textit{etc}, \textit{small}, \textit{number}] & 65.05$\pm$5.59 & 63.64$\pm$7.16 & 64.31$\pm$6.34\\
              & Sub2 [\textit{page}, \textit{topic}, \textit{value}]  & \textbf{47.46$\pm$2.18} & \textbf{53.50$\pm$1.01} & \textbf{50.26$\pm$1.16} & Sub2 [\textit{page}, \textit{non}, \textit{part}, \textit{price}] & \underline{65.57$\pm$3.04} & \underline{64.71$\pm$3.97} & \underline{65.12$\pm$3.38}\\
        \midrule
        \multirow{4}[4]{*}{7} & IRLVT [\textit{x}, \textit{y}, \textit{z}] & 50.30$\pm$2.26 & \textbf{54.14$\pm$3.48} & \underline{52.08$\pm$2.26} & IRLVT [\textit{w}, \textit{x}, \textit{y}, \textit{z}] & 66.08$\pm$3.26 & 64.73$\pm$5.08 & 65.32$\pm$3.74\\
              & ORIG [\textit{header}, \textit{question}, \textit{answer}] & \textbf{50.62$\pm$3.26} & 53.63$\pm$2.89 & 51.98$\pm$2.00 & ORIG [\textit{menu}, \textit{void}, \textit{sub}, \textit{total}] & \textbf{68.02$\pm$3.16} & \textbf{66.87$\pm$4.82} & \textbf{67.40$\pm$3.76}\\
        \cmidrule{2-9}
              & Sub1 [\textit{title}, \textit{key}, \textit{value}]  & 50.22$\pm$3.20 & 53.79$\pm$3.13 & 51.88$\pm$2.56 & Sub1 [\textit{info}, \textit{etc}, \textit{small}, \textit{number}] & \underline{67.61$\pm$4.19} & \underline{66.64$\pm$5.72} & \underline{67.08$\pm$4.72}\\
              & Sub2 [\textit{page}, \textit{topic}, \textit{value}]  & \underline{50.43$\pm$2.88} & \underline{54.03$\pm$2.71} & \textbf{52.10$\pm$2.09} & Sub2 [\textit{page}, \textit{non}, \textit{part}, \textit{price}] & 66.64$\pm$3.97 & 63.59$\pm$7.00 & 65.02$\pm$5.47\\
        \bottomrule
        \end{tabular}%
  }
}
\end{table*}

\subsection{Ablation Study}
\revise{In the ablation study, we aim at study the role of the label surface names. We introduce an ablation version, \MODEL (IRLVT), by replacing the label surface names with irrelevant tokens. We also design more different sets of words as substitutes denoted Sub1 and Sub2. The detailed substitutes are introduced in Table~\ref{tab:unk}.}

\revise{To implement the ablation study, we simply replace the word embedding of label surface names. For example, in \MODEL (Sub1) on FUNSD, we use the wording embedding of \textit{title} instead of the original \textit{header}.}

\revise{From Table~\ref{tab:unk}, we compare the performance of all the ablation models. We observe that \MODEL performs differently with distinct label semantics. In most cases, the human-designed labels can provide stronger semantic correlation with the entities than the irrelevant labels so they can further improve the performance. However, there are also drops due to improper labels. Overall, we conclude that the semantic meanings of the label surface names are useful to bridge the gap between the labels and entities.}

\subsection{Spatial Correspondence Interpretation}\label{sec:vis}

In this section, we study the ability of \MODEL to capture the spatial correspondence between certain areas and the labels. The experiment is based on the results of \MODEL on FUNSD with 7 shots. As mentioned in Section \ref{sec:mmplm}, we design unique spatial identifiers for the label surface names. The identifiers are in the same form as the spatial embeddings and \MODEL inserts the identifiers into the original spatial embedding space during sequence-to-sequence training. Ideally, the model can learn where a certain label is more likely to appear. To visualize such patterns, we compute the cosine similarity matrix $M$ of identifiers and the spatial embeddings as
$M_{ij}=\cos{(\text{SpatialID}(\tau), \text{SpatialEmb((i, j))})}$
where $(i,j)$ is the normalized coordinate pair;  $\tau \in \{\tau_1,...,\tau_t\}$. Then we plot the heatmap of the similarity matrix, where the highlight areas mean the higher similarities.

\begin{figure}[t]
\centering
\subfigure[Header]{
\includegraphics[width=0.3\linewidth]{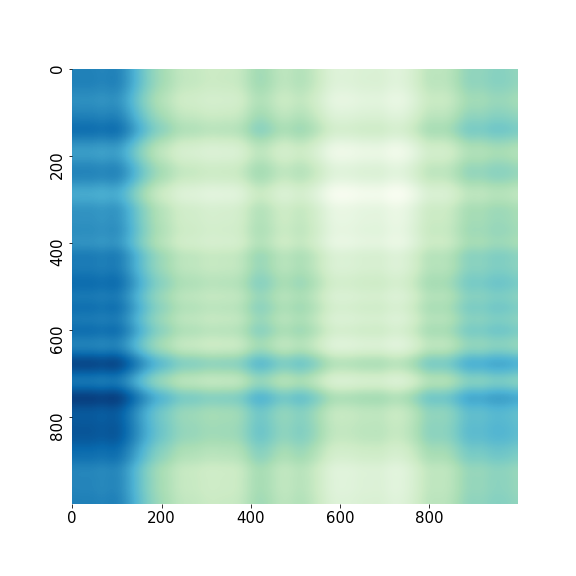}
}
\subfigure[Question]{
\includegraphics[width=0.3\linewidth]{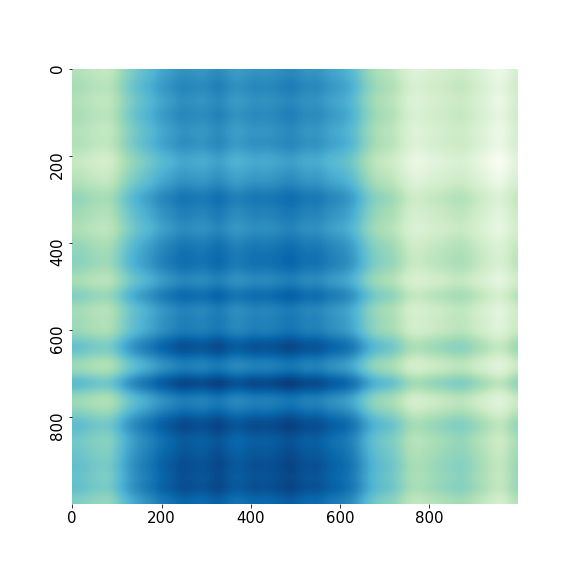}
}
\subfigure[Answer]{
\includegraphics[width=0.3\linewidth]{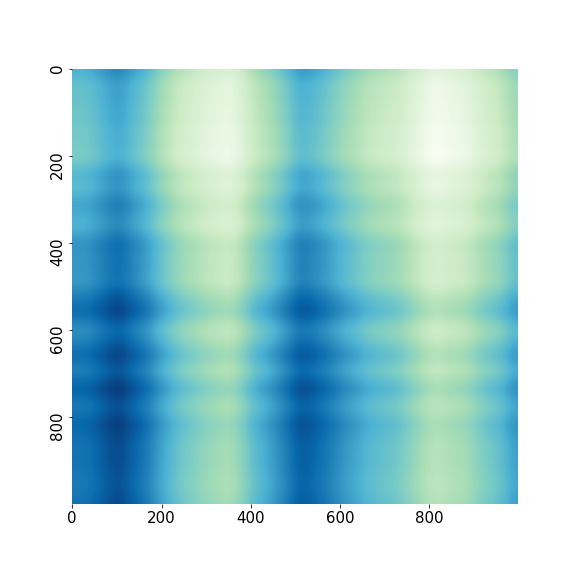}
}
\caption{Spatial correspondence visualization on FUNSD for different entity types.}
\label{fig:heatmap}
\end{figure}

\begin{figure*}[t]
\centering
\subfigure[Original Image]{
\includegraphics[width=0.2\linewidth]{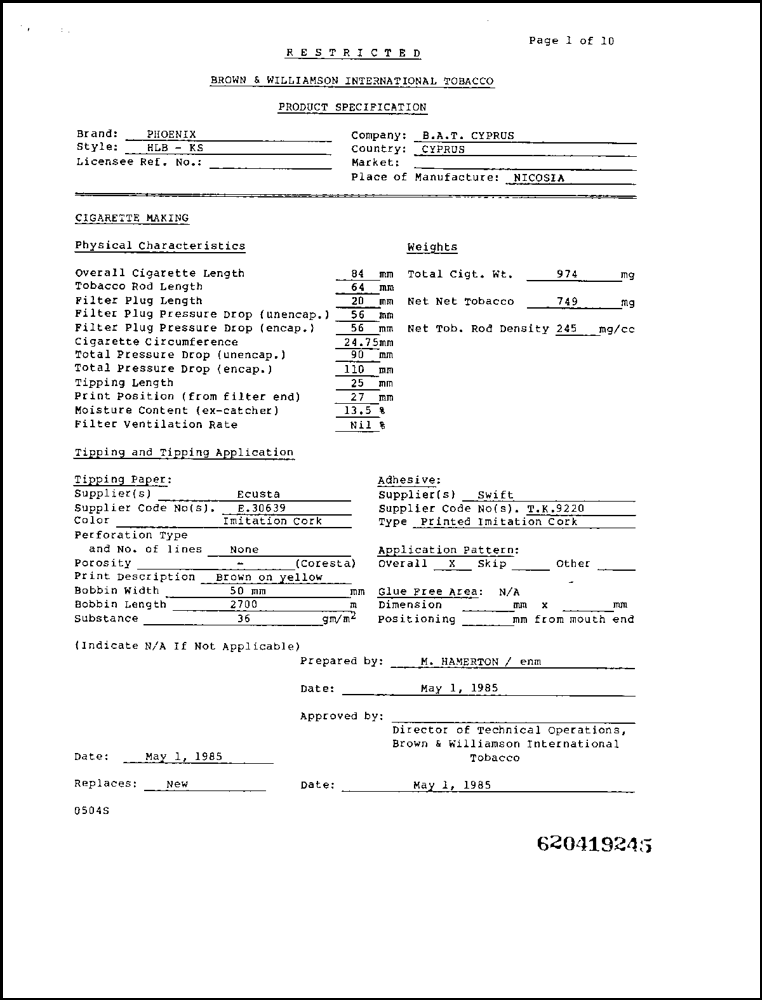}
}
\subfigure[Labeled Entities]{
\includegraphics[width=0.2\linewidth]{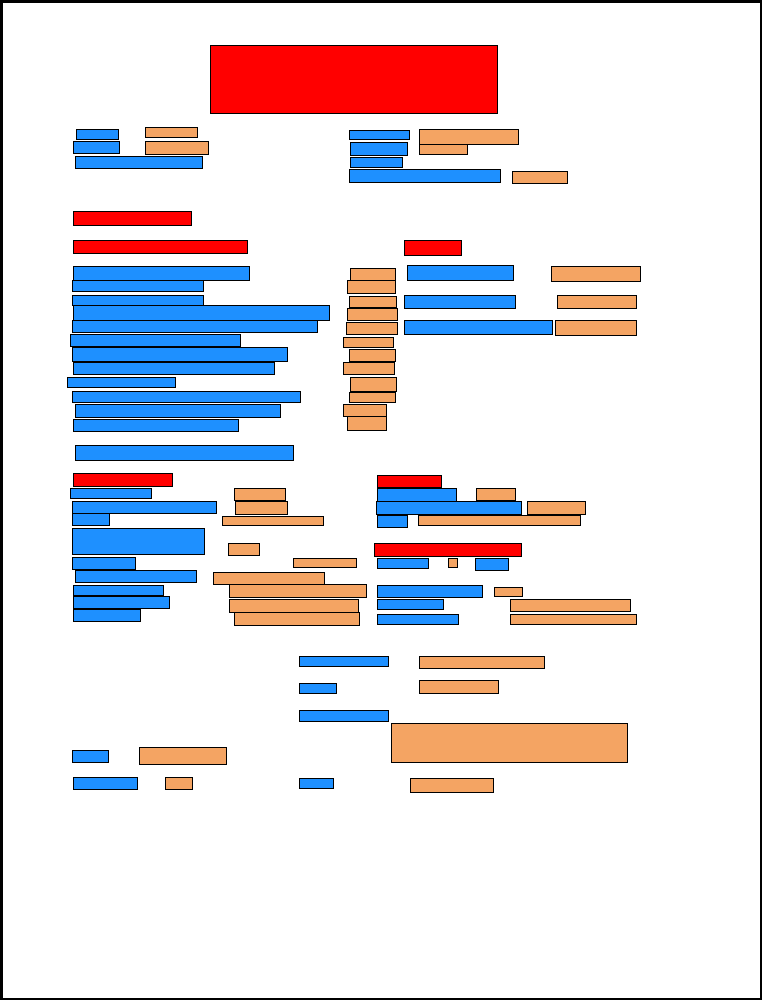}
}
\subfigure[Original Image]{
\includegraphics[width=0.2\linewidth]{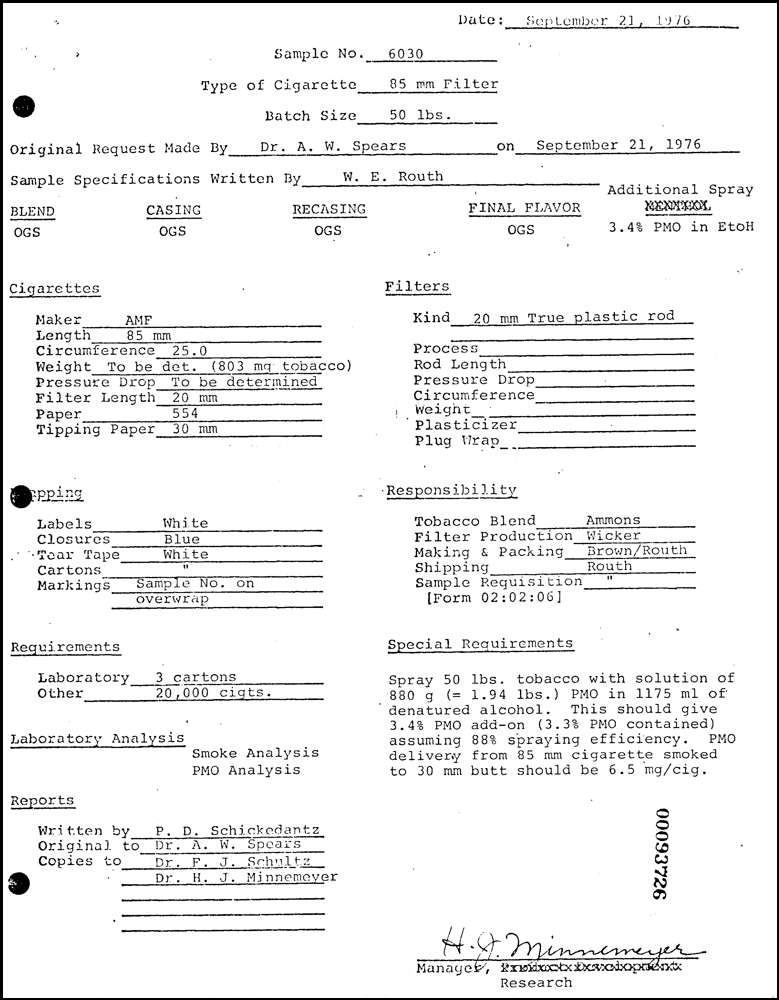}
}
\subfigure[Labeled Entities]{
\includegraphics[width=0.2\linewidth]{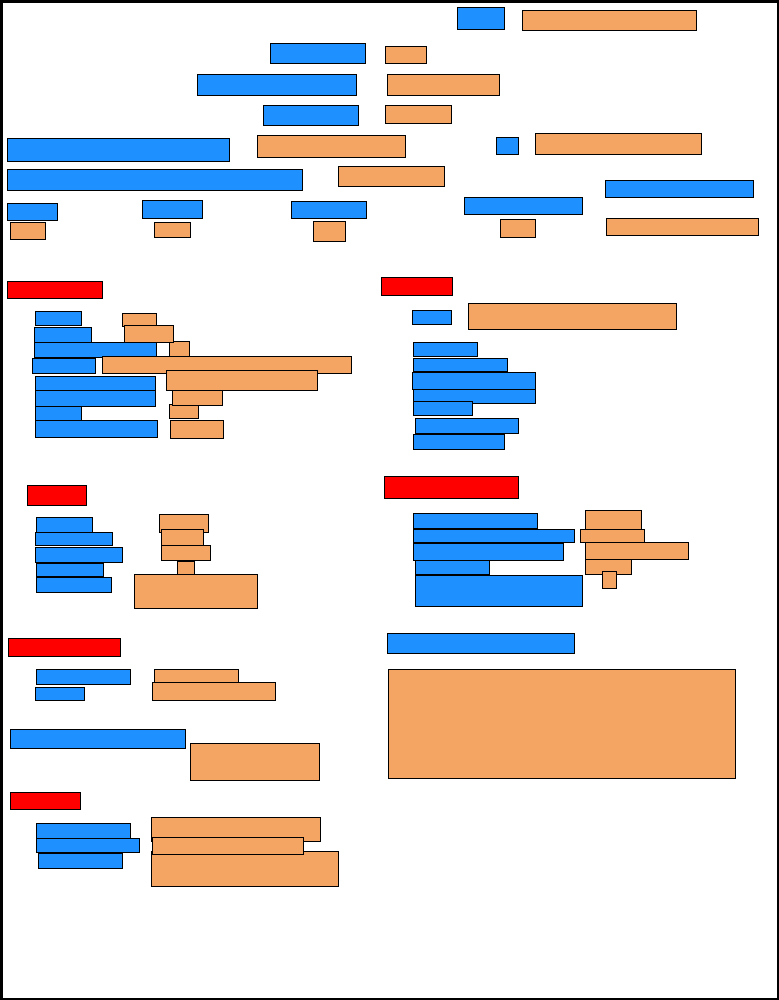}
}
\caption{Layout Format Examples from FUNSD: 
\colorbox[rgb]{0.12,0.56,1.0}{\textcolor[rgb]{0.12,0.56,1.0}{Bl}},
\colorbox[rgb]{0.96,0.64,0.38}{\textcolor[rgb]{0.96,0.64,0.38}{Bl}},
\colorbox[rgb]{1,0,0}{\textcolor[rgb]{1,0,0}{Bl}}
denotes \texttt{question}, \texttt{answer}, \texttt{header}.}
\label{fig:example}
\end{figure*}

From Figure~\ref{fig:heatmap}, we observe that the label \texttt{header} is more likely to be in the middle column of the page and may appear in the bottom part as well when there are multiple paragraphs. Intuitively, the label \texttt{question} and \texttt{answer} should appear in pairs and it is observed in Figure~\ref{fig:heatmap} that their heatmaps are almost complementary to each other. Several examples from FUNSD are selected to demonstrate the visualization results in \ref{fig:example}. Comparing the examples and the visualization results, we conclude that the spatial identifiers of labels capture the formats of pages and \MODEL leverages these features to better extract the entities under few shot settings.

\begin{figure*}[t]
\centering
\subfigure[Test Image and Expected Labels]{
\includegraphics[width=0.3\linewidth]{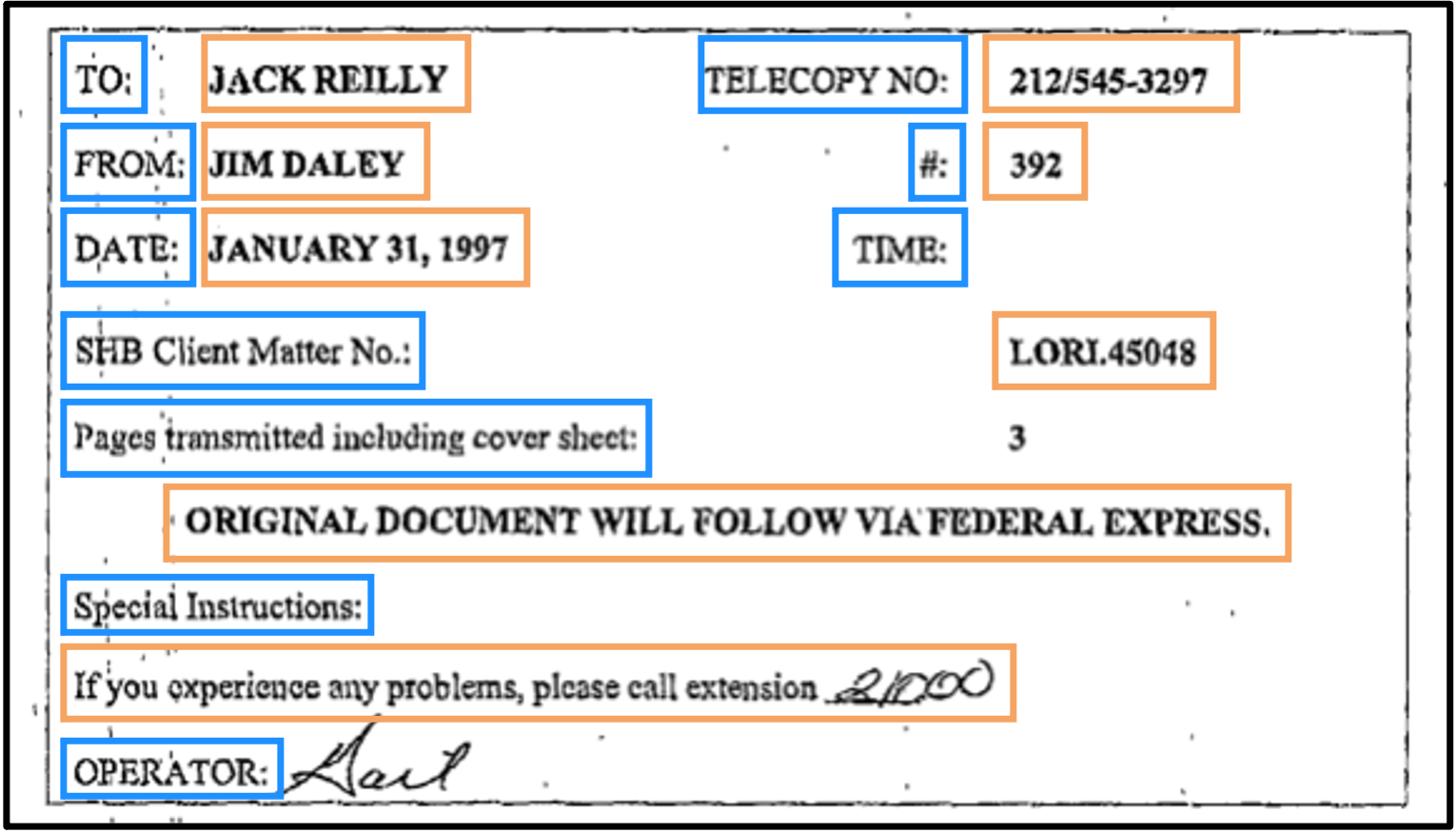}
}
\subfigure[LASER Results]{
\includegraphics[width=0.3\linewidth]{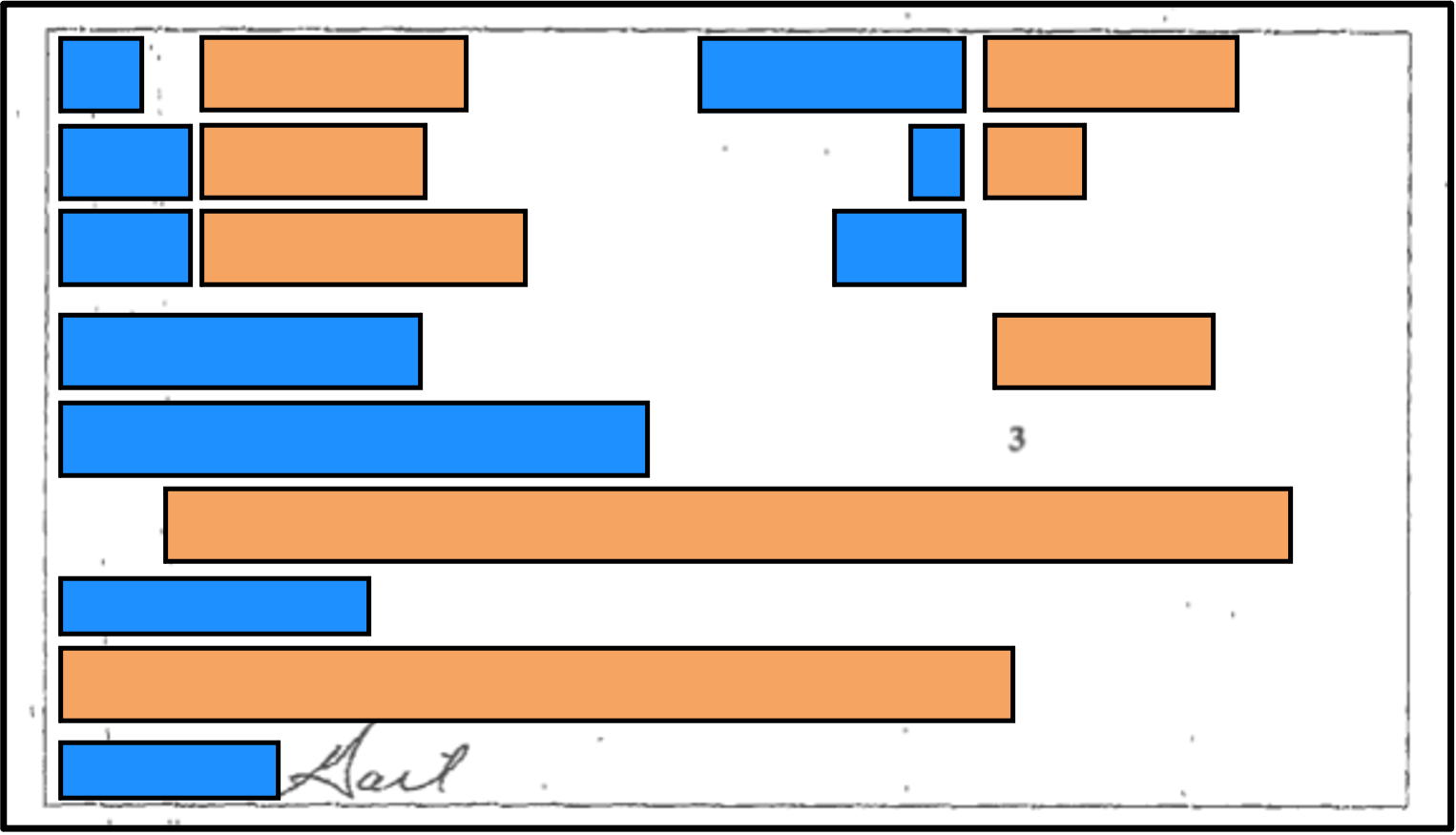}
\label{fig:funsd-laser}
}
\subfigure[LayoutLM Results]{
\includegraphics[width=0.3\linewidth]{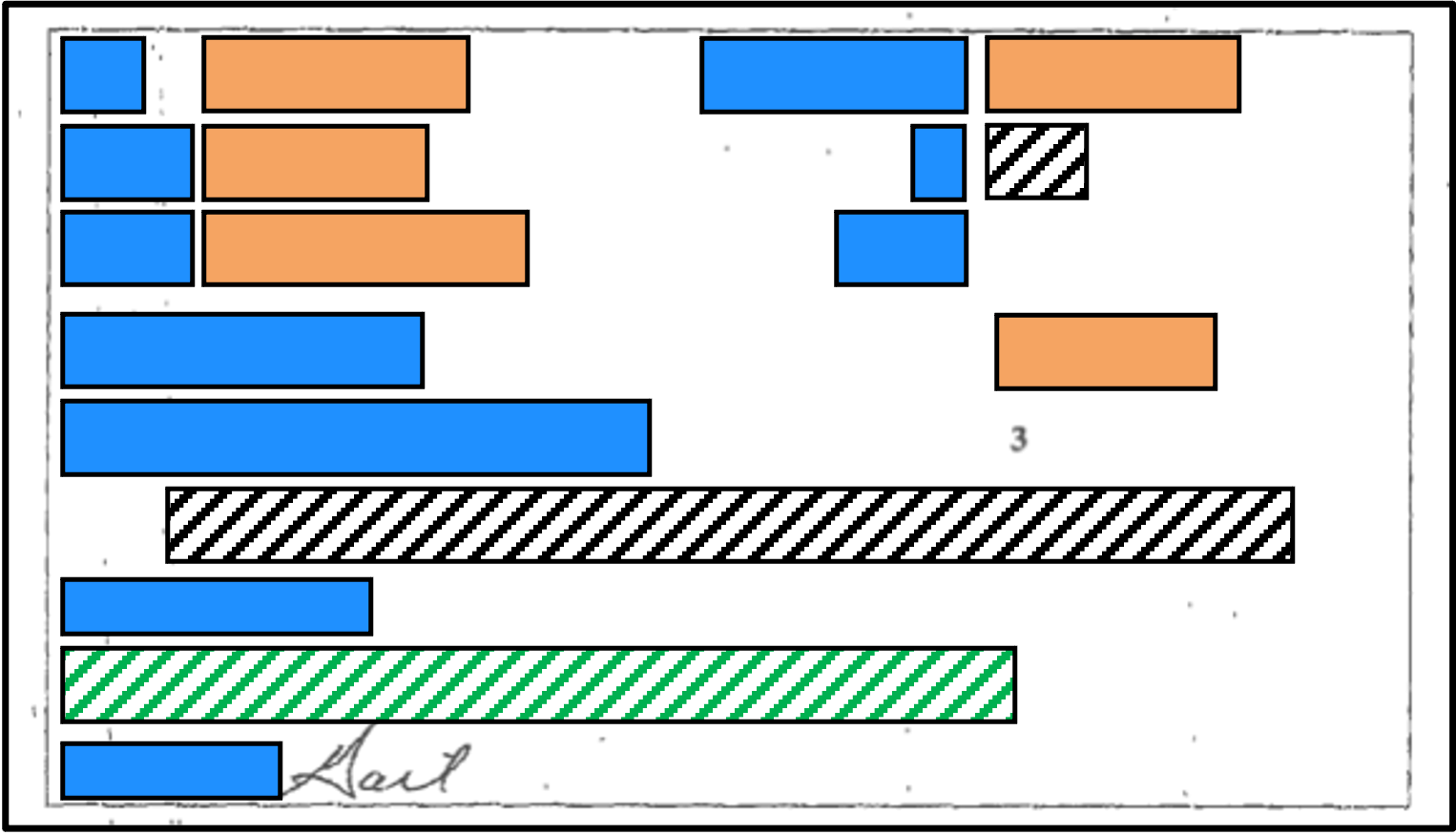}
}
\\
\subfigure[Test Image and Expected Labels]{
\includegraphics[width=0.3\linewidth]{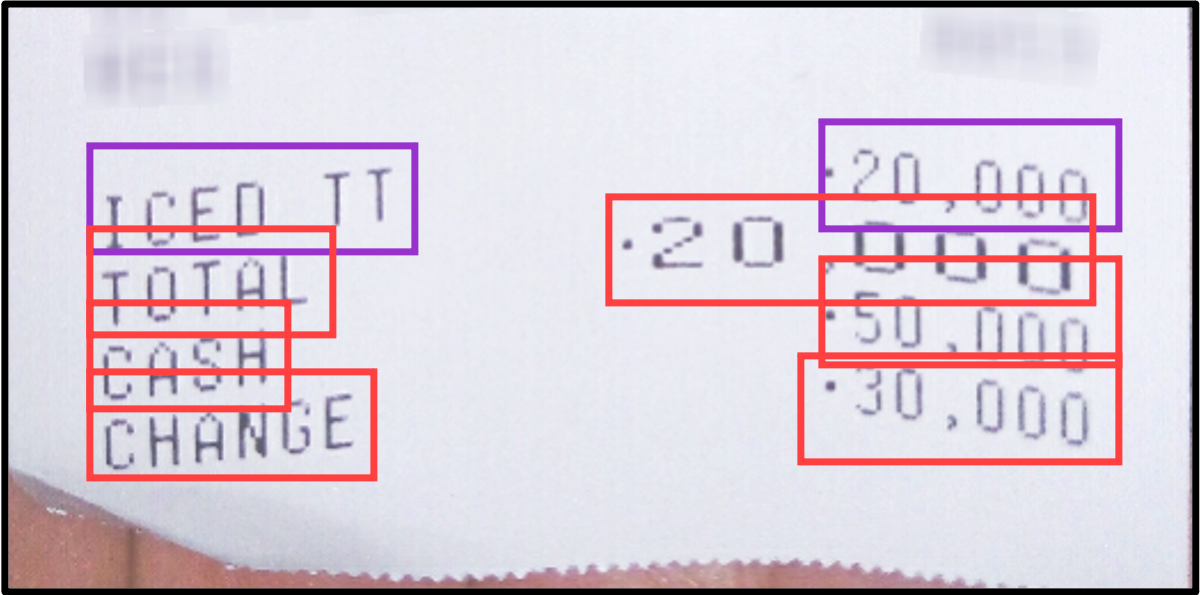}
}
\subfigure[LASER Results]{
\includegraphics[width=0.3\linewidth]{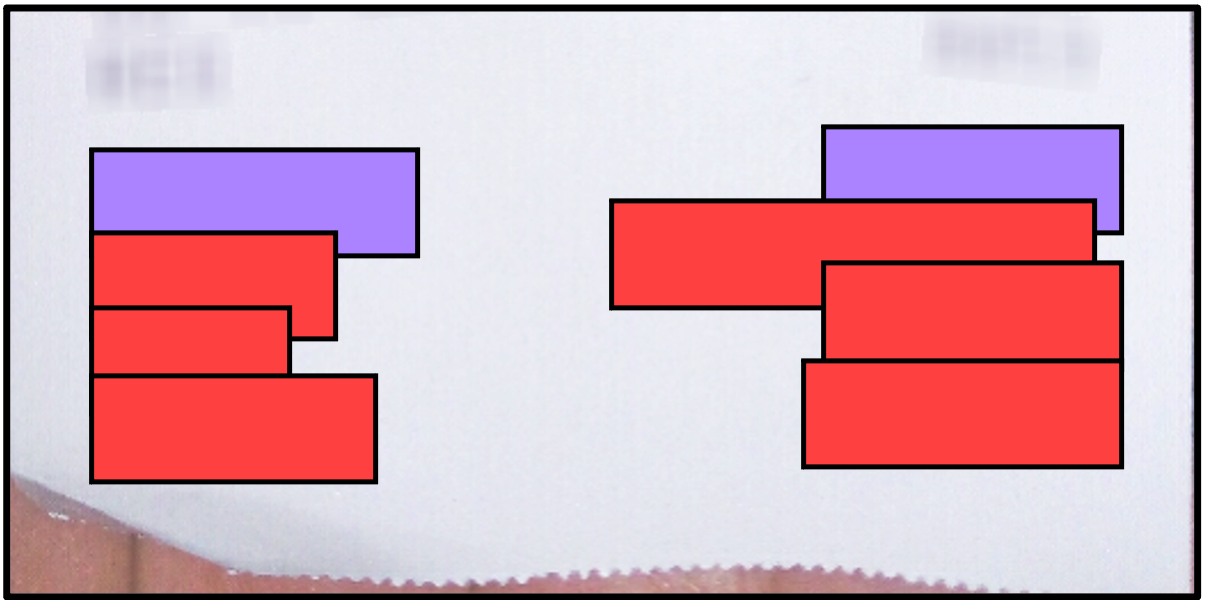}
\label{fig:cord-laser}
}
\subfigure[LayoutLM Results]{
\includegraphics[width=0.3\linewidth]{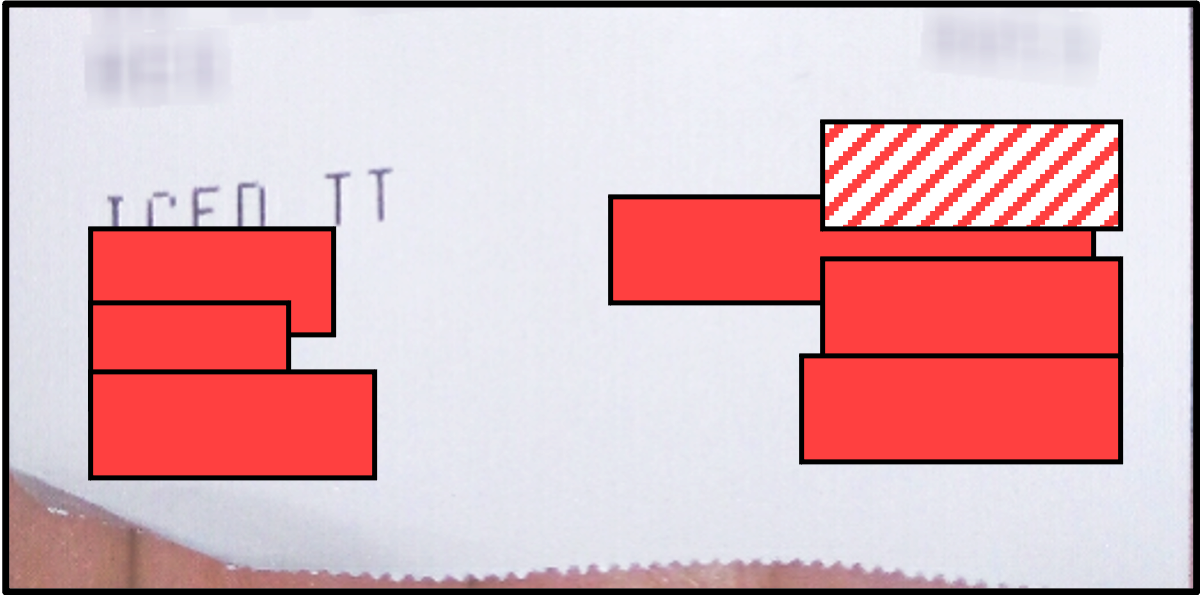}
}
\caption{Case Studies. (a), (b), (c) from FUNSD; (d), (e), (f) from CORD-Lv1; 
\colorbox[rgb]{0,0.69,0.31}{\textcolor[rgb]{0,0.69,0.31}{Bl}},
\colorbox[rgb]{0.12,0.56,1.0}{\textcolor[rgb]{0.12,0.56,1.0}{Bl}},
\colorbox[rgb]{0.96,0.64,0.38}{\textcolor[rgb]{0.96,0.64,0.38}{Bl}},
\colorbox[rgb]{0,0,0}{\textcolor[rgb]{0,0,0}{Bl}}
denote \texttt{menu}, \texttt{question}, \texttt{answer}, \texttt{other}; 
\colorbox[rgb]{0.67,0.51,1.0}{\textcolor[rgb]{0.67,0.51,1.0}{Bl}},
\colorbox[rgb]{1,0.25,0.25}{\textcolor[rgb]{1,0.25,0.25}{Bl}}
denote \texttt{menu}, \texttt{total};
\fbox{\textcolor{white}{///}}, \fbox{\textcolor{black}{///}} denote the right, wrong predictions.
}
\label{fig:case}
\end{figure*}

\subsection{Case Study}

We visualize cases from the 5-shot setting. From Figure~\ref{fig:case}, we observe \MODEL can extract the entities correctly, and the errors of LayoutLM comes from the failure to extract the entities or wrong entity type predictions. Since the sequence labeling groups the words into spans through \texttt{IOBES} tagging, which creates great uncertainty. Meanwhile, \MODEL also learns \texttt{question}s and \texttt{answer}s appear in pairs (see Figure~\ref{fig:funsd-laser}). It also
properly predicts a numerical string as \texttt{menu} even if numbers are likely to be \texttt{total} (see Figure~\ref{fig:cord-laser}).

\begin{table}[h]
  \centering
  \caption{Text-only Dataset Statistics}
  \small
\resizebox{\linewidth}{!}{
    \setlength{\tabcolsep}{3mm}{
    \begin{tabular}{lccc}
    \toprule
    \textbf{Dataset} & \textbf{\# Train} & \textbf{\# Test} & \textbf{\# Entity Type} \\
    \midrule
    OntoNotes & 60.0k & 8.3k  & 18 \\
    Mit Movie & 7.8k  & 2.0k  & 12 \\
    \bottomrule
    \end{tabular}%
    }}
  \label{tab:text-only-data}%
\end{table}%

\begin{table}[h]
  \centering
  \caption{Results of 10-way-5-shot Experiments}
  \small
\resizebox{\linewidth}{!}{
    \setlength{\tabcolsep}{4mm}{
    \begin{tabular}{lcc}
    \toprule
    \multirow{2}[4]{*}{\textbf{Model}} & \textbf{OntoNotes} & \textbf{MIT Movie} \\
    \cmidrule{2-3}  & \textbf{F-1} & \textbf{F-1} \\
    \midrule
    BERT  & 60.79$\pm$0.97 &  47.88$\pm$0.97 \\
    RoBERTa~\citep{huang2020few} & 57.70 & 51.30 \\
    UniLM &  60.82$\pm$1.26     &  51.09$\pm$1.40  \\
    \midrule
    LASER  & \textbf{61.11$\pm$1.08} & \textbf{51.88$\pm$1.27} \\
    \bottomrule
    \end{tabular}%
    }}
  \label{tab:text-only}%
\end{table}%

\subsection{Text-only Entity Recognition}

\MODEL is designed for the entity recognition task in document images where both text and layout can be leveraged to acquire essential information. However, the generative labeling scheme is not constrained in the scenario of document images. We briefly explore the potential of the generative labeling scheme in text-only scenario. We initialize \MODEL with a text-only language model, UniLM~\citep{dong2019unified}, based on the experiments in \citet{wang2021layoutreader}, and apply it onto text-only entity recognition task. Following \citet{huang2020few}, we conduct 10-way-5-shot experiments on two datasets, OntoNotes~\citep{weischedel2013ontonotes} and MIT Movie~\citep{liu2013query}, which cover general domains and review domains, respectively. The dataset statistics are shown in Table \ref{tab:text-only-data} and the results are as shown in Table \ref{tab:text-only}. We observe that our method can also surpass the sequence labeling methods in these two datasets, showing the great potential of the generative labeling scheme in the entity recognition tasks.

\section{Related Work}
\smallsection{Layout-aware LMs} Since the post-OCR processing has great application prospects, existing works propose to adapt the language pre-training to the layout formats learning. LayoutLM~\citep{xu2020layoutlm} is the pioneer in this area, which successfully uses the coordinates to represent the layout information in the embedding layer of BERT~\citep{devlin2018bert}. Following LayoutLM, the upgraded version, LayoutLMv2~\citep{Xu2020LayoutLMv2MP}, is further proposed to leverage the visual features and benefits from the alignment between words and the regions in the page. LAMBERT~\citep{garncarek2021lambert} and BROS~\citep{hong2020bros} continue studying the layout representation which uses the sinusoidal function or apply the relative positional biases from T5~\citep{raffel2019exploring}. \revise{LayoutReader~\citep{wang2021layoutreader} aims to predict the reading order of words from the OCR results. ReadingBank~\citep{wang2021layoutreader} is proposed to facilitate the pre-training of reading order detection, which annotates the reading order of millions of pages.}

\smallsection{Generalized Seq2Seq} Sequence-to-sequence architecture is basic in natural language processing and is originally designed for machine translation. With the rise of large pre-trained models, sequence-to-sequence models are increasingly used with new problem formulation. Existing works exploit the potential latent knowledge and stronger representation ability of sequence-to-sequence modeling. GENRE~\citep{de2020autoregressive} creatively reformulates the entity retrieval task into the sequence-to-sequence settings. It inferences the lined entities using the generation of BART. Recent works on prompt learning also leverage the pre-trained sequence-to-sequence language models to conduct few shot learning~\citep{liu2021pre,puri2019zero,hambardzumyan2021warp}.

\section{Conclusions and Future Work}

In this paper, we present \MODEL, a label-aware sequence-to-sequence framework for entity recognition in document images under few-shot settings. It benefits from the generative labeling scheme which reformulates the entity recognition task into the sequence-to-sequence setting. The label surface names are embedded into the generated sequence. Compared with the sequence labeling methods, \MODEL leverages the rich semantics of the label surface names and overcome the limitation of training data. Moreover, we design spatial identifiers for each label and well insert them into the spatial embedding hyperspace. In this way, \MODEL can inference the entity labels from the layout formats perspective and empirical experiments demonstrate our method can learn the layout formats though limited number of training samples. 

For further research, we will investigate the selection of label surface names and how to better leverage the semantics from the pre-trained sequence-to-sequence models. We also notice that such labeling scheme can cope with unknown categories. We will focus on the generalization of our method. 

\section*{Acknowledgments}
We want to thank the anonymous reviewers for their insightful comments. Zilong would also like to thank the Powell Fellowship for the support during the challenging years at the beginning of his Ph.D. program.

The research was sponsored in part by National Science Foundation Convergence Accelerator under award OIA-2040727 as well as generous gifts from Google, Adobe, and Teradata. Any opinions, findings, conclusions, or recommendations expressed herein are those of the authors and should not be interpreted as necessarily representing the views, either expressed or implied, of the U.S. Government. The U.S. Government is authorized to reproduce and distribute reprints for government purposes notwithstanding any copyright annotation hereon.

\section*{Ethical Consideration}
This paper focuses on the entity recognition in document images under few-shot setting. Our architecture are built upon open-source models and all the datasets are available online. We will release the code and datasets on \url{https://github.com/zlwang-cs/LASER-release}. Therefore, we do not anticipate any major ethical concerns.

\bibliography{anthology,custom}
\bibliographystyle{acl_natbib}

\clearpage\newpage
\appendix
\section*{Appendix}

\section{All Results}
All results are listed in Table~\ref{tab:everything}.
\begin{table*}[hbt]
    \centering
    \caption{Evaluation Results with Different Sizes of Few-shot Training Samples}
    \label{tab:everything}%
\small
\renewcommand{\arraystretch}{0.65}
\begin{tabular}{clcccccc}
    \toprule
    \multirow{2}[4]{*}{$|\mathcal{P}|$} & \multicolumn{1}{l}{\multirow{2}[4]{*}{\textbf{Model}}} & \multicolumn{3}{c}{\textbf{FUNSD}} & \multicolumn{3}{c}{\textbf{CORD-Lv1}} \\
    \cmidrule{3-8}      &       & \textbf{Precision} & \textbf{Recall} & \textbf{F-1} & \textbf{Precision} & \textbf{Recall} & \textbf{F-1} \\
    \midrule    
    \multirow{8}[3]{*}{1} & BERT  & 9.62$\pm$2.24 & 24.14$\pm$3.46 & 13.55$\pm$2.09 & 30.64$\pm$2.80 & 45.60$\pm$3.45 & 36.64$\pm$3.10 \\
          & RoBERTa & 9.29$\pm$1.57 & 22.06$\pm$5.64 & 12.76$\pm$1.91 & 30.66$\pm$4.25 & 44.39$\pm$6.72 & 36.25$\pm$5.18 \\
          & LayoutLM & 11.39$\pm$1.12 & 24.73$\pm$7.38 & 15.18$\pm$2.17 & 33.27$\pm$7.32 & 49.49$\pm$10.26 & 39.77$\pm$8.47 \\
          & LayoutReader & 11.32$\pm$0.62 & 22.53$\pm$4.80 & 14.84$\pm$1.25 & 32.17$\pm$4.64 & 45.61$\pm$6.54 & 37.70$\pm$5.31 \\
          & LASER & 30.40$\pm$4.89 & 35.20$\pm$7.20 & 32.36$\pm$5.14 & 47.63$\pm$3.90 & 45.52$\pm$5.84 & 46.24$\pm$3.01 \\
    \cmidrule{2-8}      & \textit{(IRLVT)} & 30.64$\pm$5.89 & 33.45$\pm$9.14 & 31.62$\pm$6.61 & 48.57$\pm$4.93 & 44.12$\pm$6.36 & 45.84$\pm$3.57 \\
          & \textit{(Sub1)} & 31.78$\pm$4.75 & 34.21$\pm$7.44 & 32.66$\pm$5.10 & 48.12$\pm$4.15 & 48.47$\pm$6.60 & 48.04$\pm$4.06 \\
          & \textit{(Sub2)} & 30.90$\pm$5.20 & 35.97$\pm$8.57 & 33.03$\pm$6.31 & 45.59$\pm$5.68 & 44.09$\pm$7.87 & 44.38$\pm$5.39 \\
    \midrule
    \multirow{8}[2]{*}{2} & BERT  & 12.49$\pm$3.24 & 30.01$\pm$4.55 & 17.53$\pm$3.89 & 37.66$\pm$3.79 & 55.43$\pm$3.41 & 44.82$\pm$3.78 \\
          & RoBERTa & 13.12$\pm$3.08 & 27.63$\pm$5.16 & 17.56$\pm$3.59 & 40.66$\pm$3.60 & 56.60$\pm$4.87 & 47.30$\pm$4.05 \\
          & LayoutLM & 19.73$\pm$5.43 & 37.32$\pm$9.58 & 25.41$\pm$6.25 & 42.92$\pm$8.27 & 60.47$\pm$8.15 & 50.15$\pm$8.44 \\
          & LayoutReader & 16.66$\pm$2.58 & 29.26$\pm$5.34 & 20.96$\pm$3.01 & 44.14$\pm$7.77 & 58.50$\pm$8.32 & 50.26$\pm$8.04 \\
          & LASER & 39.23$\pm$3.09 & 42.43$\pm$6.23 & 40.38$\pm$2.71 & 59.31$\pm$2.31 & 54.85$\pm$6.19 & 56.80$\pm$3.46 \\
\cmidrule{2-8}          & \textit{(IRLVT)} & 38.75$\pm$3.74 & 40.37$\pm$8.17 & 39.19$\pm$4.84 & 57.05$\pm$2.87 & 53.49$\pm$7.01 & 55.01$\pm$4.19 \\
          & \textit{(Sub1)} & 38.47$\pm$2.95 & 41.11$\pm$7.03 & 39.43$\pm$3.83 & 57.46$\pm$4.22 & 54.72$\pm$5.87 & 55.98$\pm$4.68 \\
          & \textit{(Sub2)} & 37.52$\pm$2.22 & 43.23$\pm$6.48 & 39.90$\pm$3.12 & 57.03$\pm$2.80 & 55.75$\pm$5.48 & 56.28$\pm$3.48 \\
    \midrule
    \multirow{8}[4]{*}{3} & BERT  & 16.42$\pm$4.30 & 34.74$\pm$5.36 & 22.19$\pm$5.05 & 39.62$\pm$3.99 & 56.65$\pm$4.03 & 46.58$\pm$3.94 \\
          & RoBERTa & 16.71$\pm$3.63 & 31.28$\pm$3.55 & 21.66$\pm$3.84 & 44.51$\pm$4.69 & 60.18$\pm$4.69 & 51.15$\pm$4.70 \\
          & LayoutLM & 28.67$\pm$6.56 & 47.22$\pm$8.31 & 35.42$\pm$7.00 & 47.68$\pm$7.49 & 63.93$\pm$7.04 & 54.57$\pm$7.46 \\
          & LayoutReader & 22.37$\pm$2.03 & 35.19$\pm$4.97 & 27.19$\pm$2.56 & 43.85$\pm$4.72 & 56.90$\pm$2.47 & 49.47$\pm$3.95 \\
          & LASER & 43.66$\pm$1.97 & 47.08$\pm$5.72 & 45.21$\pm$3.74 & 61.16$\pm$3.11 & 60.33$\pm$5.65 & 60.63$\pm$4.00 \\
    \cmidrule{2-8}      & \textit{(IRLVT)} & 43.51$\pm$1.46 & 47.92$\pm$5.93 & 45.44$\pm$3.36 & 61.50$\pm$2.52 & 59.17$\pm$4.11 & 60.27$\pm$2.99 \\
          & \textit{(Sub1)} & 43.87$\pm$1.33 & 47.11$\pm$6.07 & 45.26$\pm$3.44 & 61.54$\pm$2.76 & 58.79$\pm$6.76 & 60.00$\pm$4.57 \\
          & \textit{(Sub2)} & 43.88$\pm$1.34 & 48.01$\pm$6.86 & 45.65$\pm$3.93 & 61.85$\pm$2.16 & 60.29$\pm$2.85 & 61.03$\pm$2.10 \\
    \midrule
    \multirow{8}[2]{*}{4} & BERT  & 18.25$\pm$3.30 & 37.90$\pm$2.93 & 24.55$\pm$3.59 & 43.94$\pm$4.14 & 61.13$\pm$4.18 & 51.09$\pm$4.12 \\
          & RoBERTa & 17.99$\pm$2.84 & 34.38$\pm$4.09 & 23.52$\pm$3.07 & 49.56$\pm$4.89 & 65.19$\pm$4.85 & 56.29$\pm$4.86 \\
          & LayoutLM & 33.38$\pm$3.62 & 53.71$\pm$3.24 & 41.00$\pm$2.71 & 52.15$\pm$7.90 & 68.06$\pm$6.86 & 58.99$\pm$7.66 \\
          & LayoutReader & 24.61$\pm$2.58 & 38.28$\pm$3.05 & 29.89$\pm$2.51 & 48.27$\pm$9.14 & 61.29$\pm$8.49 & 53.96$\pm$9.06 \\
          & LASER & 44.91$\pm$2.42 & 50.25$\pm$3.26 & 47.36$\pm$2.18 & 63.90$\pm$3.19 & 60.99$\pm$5.86 & 62.38$\pm$4.60 \\
  \cmidrule{2-8} & \textit{(IRLVT)} & 46.31$\pm$1.91 & 51.74$\pm$2.55 & 48.83$\pm$1.67 & 63.68$\pm$3.72 & 60.39$\pm$7.11 & 61.93$\pm$5.50 \\
          & \textit{(Sub1)} & 45.58$\pm$1.63 & 51.47$\pm$2.97 & 48.29$\pm$1.65 & 63.22$\pm$3.31 & 60.39$\pm$8.23 & 61.65$\pm$5.93 \\
          & \textit{(Sub2)} & 45.43$\pm$2.08 & 51.74$\pm$2.64 & 48.33$\pm$1.79 & 62.85$\pm$3.83 & 60.07$\pm$8.00 & 61.31$\pm$5.97 \\
    \midrule
    \multirow{8}[4]{*}{5} & BERT  & 20.57$\pm$2.59 & 39.25$\pm$1.10 & 26.93$\pm$2.46 & 45.73$\pm$4.31 & 63.29$\pm$3.68 & 53.06$\pm$4.14 \\
          & RoBERTa & 19.47$\pm$2.32 & 35.04$\pm$1.89 & 24.94$\pm$1.93 & 52.21$\pm$4.55 & 66.63$\pm$5.52 & 58.54$\pm$4.92 \\
          & LayoutLM & 39.24$\pm$4.33 & 58.20$\pm$2.45 & 46.72$\pm$3.12 & 56.13$\pm$7.39 & 71.66$\pm$6.13 & 62.91$\pm$7.04 \\
          & LayoutReader & 27.52$\pm$3.44 & 41.17$\pm$4.01 & 32.89$\pm$3.28 & 51.97$\pm$8.42 & 63.82$\pm$7.87 & 57.24$\pm$8.32 \\
          & LASER & 47.25$\pm$1.93 & 52.85$\pm$1.22 & 49.87$\pm$1.29 & 65.62$\pm$3.79 & 64.90$\pm$5.78 & 65.23$\pm$4.70 \\
    \cmidrule{2-8}      & \textit{(IRLVT)} & 46.94$\pm$1.87 & 52.96$\pm$2.03 & 49.74$\pm$1.63 & 63.67$\pm$3.82 & 61.10$\pm$5.21 & 62.33$\pm$4.48 \\
          & \textit{(Sub1)} & 47.43$\pm$2.29 & 52.19$\pm$2.09 & 49.68$\pm$1.98 & 65.05$\pm$5.59 & 63.64$\pm$7.16 & 64.31$\pm$6.34 \\
          & \textit{(Sub2)} & 47.46$\pm$2.18 & 53.50$\pm$1.01 & 50.26$\pm$1.16 & 65.57$\pm$3.04 & 64.71$\pm$3.97 & 65.12$\pm$3.38 \\
    \midrule
    \multirow{8}[4]{*}{6} & BERT  & 20.37$\pm$2.08 & 39.58$\pm$2.90 & 26.85$\pm$2.27 & 46.39$\pm$4.56 & 63.61$\pm$4.81 & 53.62$\pm$4.65 \\
          & RoBERTa & 21.70$\pm$2.37 & 37.14$\pm$1.34 & 27.32$\pm$1.96 & 51.80$\pm$5.60 & 66.54$\pm$5.87 & 58.25$\pm$5.77 \\
          & LayoutLM & 41.75$\pm$3.83 & 60.47$\pm$3.23 & 49.28$\pm$3.02 & 59.31$\pm$4.23 & 74.48$\pm$3.16 & 66.02$\pm$3.86 \\
          & LayoutReader & 29.19$\pm$2.10 & 43.21$\pm$2.32 & 34.82$\pm$2.09 & 52.76$\pm$4.45 & 65.62$\pm$3.47 & 58.45$\pm$3.99 \\
          & LASER & 48.64$\pm$2.14 & 53.54$\pm$2.10 & 50.96$\pm$1.95 & 66.85$\pm$3.88 & 65.97$\pm$5.19 & 66.38$\pm$4.38 \\
    \cmidrule{2-8}      & \textit{(IRLVT)} & 48.33$\pm$2.15 & 52.68$\pm$1.56 & 50.36$\pm$1.08 & 66.01$\pm$4.07 & 63.85$\pm$5.59 & 64.87$\pm$4.63 \\
          & \textit{(Sub1)} & 48.49$\pm$1.96 & 52.85$\pm$1.67 & 50.53$\pm$0.99 & 65.19$\pm$4.22 & 63.37$\pm$7.10 & 64.21$\pm$5.68 \\
          & \textit{(Sub2)} & 49.14$\pm$1.97 & 53.42$\pm$1.20 & 51.16$\pm$1.22 & 66.44$\pm$3.11 & 64.05$\pm$4.31 & 65.21$\pm$3.67 \\
    \midrule
    \multirow{8}[4]{*}{7} & BERT  & 21.44$\pm$2.07 & 40.87$\pm$3.79 & 28.09$\pm$2.48 & 50.13$\pm$4.35 & 66.67$\pm$3.67 & 57.20$\pm$4.07 \\
          & RoBERTa & 23.68$\pm$3.06 & 38.74$\pm$3.54 & 29.32$\pm$3.08 & 55.14$\pm$4.49 & 69.35$\pm$4.16 & 61.43$\pm$4.42 \\
          & LayoutLM & 43.23$\pm$5.27 & 61.73$\pm$5.97 & 50.76$\pm$5.30 & 62.87$\pm$3.98 & 76.38$\pm$2.72 & 68.96$\pm$3.49 \\
          & LayoutReader & 31.22$\pm$3.14 & 45.08$\pm$3.83 & 36.85$\pm$3.26 & 54.43$\pm$5.89 & 65.48$\pm$5.34 & 59.42$\pm$5.68 \\
          & LASER & 50.62$\pm$3.26 & 53.63$\pm$2.89 & 51.98$\pm$2.00 & 68.02$\pm$3.16 & 66.87$\pm$4.82 & 67.40$\pm$3.76 \\
    \cmidrule{2-8}      & \textit{(IRLVT)} & 50.30$\pm$2.26 & 54.14$\pm$3.48 & 52.08$\pm$2.26 & 66.08$\pm$3.26 & 64.73$\pm$5.08 & 65.32$\pm$3.74 \\
          & \textit{(Sub1)} & 50.22$\pm$3.20 & 53.79$\pm$3.13 & 51.88$\pm$2.56 & 67.61$\pm$4.19 & 66.64$\pm$5.72 & 67.08$\pm$4.72 \\
          & \textit{(Sub2)} & 50.43$\pm$2.88 & 54.03$\pm$2.71 & 52.10$\pm$2.09 & 66.64$\pm$3.97 & 63.59$\pm$7.00 & 65.02$\pm$5.47 \\
    \bottomrule
    \end{tabular}%
\end{table*}%




\end{document}